\pdfoutput=1

\documentclass[11pt]{article}

\usepackage[preprint]{acl}
\usepackage{graphicx} 
\usepackage{booktabs}
\usepackage{rotating}
\usepackage{amsmath}

\usepackage{multirow}

\usepackage{times}
\usepackage{latexsym}

\usepackage[T1]{fontenc}

\usepackage[utf8]{inputenc}

\usepackage{microtype}

\usepackage{inconsolata}

\title{Quantization and Machine Translation: \\When Do LLMs Forget Languages?}
\title{How Do Quantization Methods, Bit-Widths, and Languages, \\Impact Machine Translation with LLMs}
\title{Impact of Quantization on Machine
Translation with Large Language Models: \\Investigation into Bit-Widths and Languages}
\title{The Uneven Impact of Post-Training Quantization in Machine Translation}
\author{
  \textbf{Benjamin Marie} and 
  \textbf{Atsushi Fujita\textsuperscript{1}}
\\
\\
  \textsuperscript{1}NICT
\\
  \small{
    \textbf{Correspondence (first author):} {contact@kaitchup.com}
  }
}

\begin{document}
\maketitle
\begin{abstract}
Quantization is essential for deploying large language models (LLMs) on resource-constrained hardware, but its implications for multilingual tasks remain underexplored. We conduct the first large-scale evaluation of post-training quantization (PTQ) on machine translation across 55 languages using five LLMs ranging from 1.7B to 70B parameters. Our analysis reveals that while 4-bit quantization often preserves translation quality for high-resource languages and large models, significant degradation occurs for low-resource and typologically diverse languages, particularly in 2-bit settings. We compare four quantization techniques (AWQ, BitsAndBytes, GGUF, and AutoRound), showing that algorithm choice and model size jointly determine robustness. GGUF variants provide the most consistent performance, even at 2-bit precision. Additionally, we quantify the interactions between quantization, decoding hyperparameters, and calibration languages, finding that language-matched calibration offers benefits primarily in low-bit scenarios. Our findings offer actionable insights for deploying multilingual LLMs for machine translation under quantization constraints, especially in low-resource settings.
\end{abstract}

\section{Introduction}
\begingroup
\renewcommand\thefootnote{}\footnote{\noindent This work was partly conducted by the first author during a research visit to NICT.}
\addtocounter{footnote}{-1}
\endgroup
Quantization is now a key technique for scaling large language models (LLMs) to real-world deployments, enabling substantial reductions in memory usage and computational overhead, particularly on consumer-grade hardware. However, most existing research on quantization focuses exclusively on English language models and tasks, overlooking the linguistic diversity that multilingual LLMs are more and more often trained to support \cite{gemmateam2025gemma3technicalreport,yang2025qwen3technicalreport}. In practice, multilingual applications, especially machine translation (MT), are often deployed in low-resource settings, where both language data and computing capacity are limited. Ensuring that LLMs remain robust and accurate under quantization is thus essential for inclusive and globally accessible AI.

Despite its importance, the impact of quantization on MT remains underexplored. Prior work \cite{ogueji-etal-2022-intriguing,marchisio-etal-2024-quantization} suggests that compression techniques such as quantization and sparsity can disproportionately affect long-tail linguistic features, often corresponding to underrepresented languages. However, a comprehensive evaluation spanning languages, model scales, and quantization methods has yet to be conducted.

In this study, we address this gap by systematically evaluating a suite of LLMs on MT tasks under various quantization settings. We consider five models ranging from 1.5B to 70B parameters, apply multiple quantization algorithms, including low-bit formats such as 2-bit and 4-bit, and assess translation performance across 55 language pairs, covering both low- and high-resource languages.

By focusing on the intersection of MT, multilinguality, and quantization, our work provides the first detailed analysis of how quantized LLMs perform across diverse languages and quantization strategies. Our key findings are as follows:
\begin{itemize}
\item Quantization retains strong performance in MT, even at low bitwidths, for the larger models, and most language pairs. Smaller models (<10B parameters) show clear signs of degradation in MT, especially for low-resource languages.

\item Qwen3 and Llama 3.1 models demonstrate some ability to translate languages outside their official training data. However, quantization substantially degrades performance for these lower-resource languages, especially Indic languages, and in some cases, eliminates their ability to translate them into English.

\item Quantization method matters: A calibration-free method like BitsAndBytes (bnb) performs well for smaller models in MT but shows limitations at larger scales. Surprisingly, simple formats like GGUF Q2\_K yield strong performance, even for sub-10B parameter models.

\item  We dissect decoding temperature, nucleus sampling, and calibration-set language, showing that (i) sampling temperature dominates top-$p$; (ii) calibration language helps only at extreme (2-bit) quantization; and (iii) these factors interact additively with weight precision, yielding simple heuristics for real-world MT deployment on commodity hardware.

\end{itemize}

\section{Background: Post-Training Quantization}

Quantization is a technique for reducing the computational and memory
footprint of neural networks by representing weights and optionally
activations in lower-bit formats. This can be done during training,
known as quantization-aware training (QAT), or applied post hoc to a
pre-trained model, known as post-training quantization (PTQ). For LLMs,
training directly at reduced precision, especially below 16-bit, presents
significant optimization and engineering challenges.
Consequently, PTQ is often preferred in practice, as it allows models to be
trained at high precision and then quantized in a single pass without further gradient updates.

In this work, we focus on weight-only PTQ due to its simplicity and
scalability. Only the weight matrices are quantized, while activations and
the KV cache remain in higher precision. 

A standard symmetric uniform quantization scheme maps a real-valued weight
tensor $w$ to a quantized integer tensor\footnote{INT data types are the most used for quantization as they are optimized for computation with modern GPUs. Other data types used for quantization can be FP8 and FP4
variants.} $w_q$ via
\[
w_q = \mathrm{round}\left( \frac{w}{s} \right), \quad \hat{w} = s \cdot w_q
\]
where $s$ is the scale factor that determines the resolution of
quantization. A common choice for $s$ in symmetric quantization is
\[
s = \frac{\max(|w|)}{2^{N - 1} - 1}
\]
where $N$ is the bitwidth of quantization (e.g., 8 or 4), and the
denominator corresponds to the maximum representable integer value in the
signed range. The rounding operation ensures that values are mapped to the
nearest quantized level, and the multiplication during dequantization
approximately reconstructs the original value. In practice, $w_q$ is
clamped to the integer range $[-2^{N-1} + 1, 2^{N-1} - 1]$ so the negative and positive halves are perfectly symmetric, although many quantization implementations handle this implicitly.

Using a single global scale factor for an entire weight tensor (per-tensor quantization) is computationally efficient. Still, it may introduce significant errors when the distribution of weights varies widely across channels.\footnote{A channel is a chosen tensor axis (e.g., each row/column of a weight matrix or an attention head) along which distinct quantization parameters (scale/zero-point) are applied in per-channel schemes.} As a common refinement to reduce quantization errors in low-bit
settings (e.g., 4-bit or lower), group-wise quantization is often employed, in which input channels are divided into fixed-size groups (e.g., 128), and each group is assigned its own scale. These more granular approaches improve approximation quality while maintaining the benefits of reduced precision.

Although uniform quantization is effective at moderate bitwidths, it often leads to significant degradation at low precision. Previous works, such as GPTQ \cite{frantar2023gptqaccurateposttrainingquantization} and AWQ \cite{MLSYS2024_42a452cb}, incorporate
quantization-aware error minimization during the post-training process, leading to substantial improvements in accuracy for quantized LLMs. These methods extend basic weight-only PTQ by optimizing the quantization step to preserve downstream performance more effectively.

\section{Related Work}

A growing body of research on compression techniques, and particularly quantization, explores multilingual settings. However, most studies exhibit one of two key limitations: they lack core multilingual task evaluations or fall back on English-centric benchmarks.

\citet{paglieri2024outlierscalibrationsetsdiminishing}, for instance, introduce multilingual calibration datasets but limit their analysis to English, leaving multilingual generalization untested. \citet{kharazmi-icassp2023} demonstrate that compression-induced degradation in LSTMs is significantly more severe in multilingual contexts compared to monolingual scenarios, suggesting that language diversity intensifies performance losses.

MT is a critical metric of multilingual capability, yet it is often omitted from compression research. When translation is evaluated, results are mixed: \citet{diddee-etal-2022-brittle} show that effectiveness of knowledge distillation varies by language, influenced heavily by teacher model confidence and the use of synthetic data.

Multilingual corpora exemplify the long-tail problem in NLP, where low-resource or typologically rare languages behave like rare features during training. Several studies, \citet{hooker2020characterisingbiascompressedmodels, ahia-etal-2021-low-resource, hooker2021compresseddeepneuralnetworks,ogueji-etal-2022-intriguing}, report that compression disproportionately harms such features or languages. Yet compression is not universally detrimental: under certain scenarios, quantization or sparsity can act as a form of implicit regularization, improving downstream performance, though generally favoring high-frequency elements (\citealt{ahia-etal-2021-low-resource}; \citealt{ogueji-etal-2022-intriguing}; \citealt{anonymous2024quantifying};).

Recent advances in quantizing LLMs, NormalFloat4 (\citealt{10.5555/3666122.3666563}), LLM.int8 (\citealt{10.5555/3600270.3602468}), and structured methods (\citealt{ahmadian2023intriguing,bondarenko2023quantizable,10.5555/3618408.3619993, 10.1609/aaai.v38i16.29765}), achieve impressive efficiency gains. Still, their evaluation remains monolingual. The comprehensive comparison of bitwidths recommends 4-bit as an optimal trade-off, but is conducted exclusively in English (\citealt{10.5555/3618408.3618715}). Later work on quantized Llama 3 (\citealt{Huang2024}) and size-precision trade-offs (\citealt{anonymous2024quantifying}) similarly omit multilingual analysis.

\citet{marchisio-etal-2024-quantization} present the first in-depth evaluation of quantization on multilingual LLMs, covering models up to 103B parameters and 23 languages. They employ a rich mix of automatic benchmarks (e.g., mMMLU, MGSM, FLORES), ``Language Confusion'' tests, LLM-as-a-Judge, and human evaluations. Their results underscore the need for multilingual-aware evaluation in model compression. Their study mainly tackles high-resource languages, large models, and does not involve low-bit quantization.

To the best of our knowledge, our work is the first to systematically evaluate LLMs quantized with different bit-widths and techniques, specifically for MT tasks involving high and low-resource languages.

\section{Experiments}
\subsection{Evaluation Task and Inference Settings}
We evaluated all models using WMT24++\footnote{\url{https://huggingface.co/datasets/google/wmt24pp}} \cite{deutsch-etal-2025-wmt24}, a multilingual benchmark covering 55 languages and four domains (literary, news, social, and speech). We conducted translation evaluations in all directions to and from English. Evaluation was conducted with COMET \cite{rei-etal-2022-comet} (specifically, \texttt{wmt22-comet-da}\footnote{\url{https://huggingface.co/Unbabel/wmt22-comet-da}}). While COMET is effective for system ranking, the absolute scores are not interpretable and can be misleading, e.g., yielding high scores for outputs that are not translations, empty, or in the source language \citep{perrella-etal-2024-guardians}.

Given the 110 translation directions\footnote{We conduct experiments in both En$\rightarrow$Xx and Xx$\rightarrow$En directions. However, it is important to note that WMT24++ was originally constructed for translation from English into other languages. As a result, the reverse direction (Xx$\rightarrow$En) creates an artificial scenario in which the source text may exhibit translationese, i.e., language that tends to be simpler and more regular than the original text, making it potentially easier to translate.} and the fact that metric scores cannot be meaningfully averaged across language pairs,\footnote{Averaging BLEU \cite{papineni-etal-2002-bleu}, chrF \cite{popovic-2015-chrf}, or COMET scores across translation directions is a common practice in MT but is theoretically flawed and methodologically unsound. These metrics lack interval-scale properties and are not calibrated across languages, making direct comparisons invalid. BLEU and chrF, for instance, penalize morphologically rich or low-resource languages disproportionately \cite{freitag-etal-2021-experts, popovic-2015-chrf}. Even COMET, while learned, is reliable only for relative comparisons within a specific language pair \cite{rei-etal-2020-comet,zouhar-etal-2024-pitfalls}. As \citet{graham-etal-2014-machine} and \citet{graham-baldwin-2014-testing} argue, averaging obscures disparities, inflates scores for high-resource languages, and reduces correlation with human judgment. Accurate multilingual evaluation requires either per-language reporting or rank-based comparisons.} we report results for a representative subset of six languages, selected for their script diversity and their varying representation in LLM training corpora: Japanese (ja\_JP), French (fr\_FR), Polish (pl\_PL), Bengali (bn\_IN), Malayalam (ml\_IN), and Zulu (zu\_ZA).\footnote{Language-region codes follow those used in the WMT24++ dataset.} The full translations for the 110 translation directions can be found in this dataset repository: bnjmnmarie/wmt24pp-qtranslated.\footnote{\url{https://huggingface.co/datasets/bnjmnmarie/wmt24pp-qtranslated}}

As our focus is on measuring the relative degradation in translation quality due to quantization, we used the same simple translation prompt for all models.\footnote{We assume that task performance could be improved through prompt tuning, few-shot learning, or model-specific prompting strategies, but we opted for a uniform setup to isolate quantization effects, while using each model's chat template.}
\subsection{Models}
For these experiments, given the correlation between quantization accuracy and model size, we selected five models of varying scales from the Qwen3 \cite{yang2025qwen3technicalreport}\footnote{Qwen3 are reasoning models whose reasoning mode can be disabled. We disabled it for all experiments.} and Llama 3 \cite{grattafiori2024llama3herdmodels} families: Qwen3-1.7B, Qwen3-8B, Llama-3.1-8B-Instruct, Qwen3-32B, and Llama-3.3-70B. All models are primarily designed for English-language, in addition to Chinese for Qwen3, tasks.\footnote{We also conducted preliminary experiments with Llama-3.1-Swallow-70B-Instruct-v0.3 \cite{fujii2024continual}, which is based on Llama-3.1-70B and further trained via continual pretraining on 200B Japanese tokens. We aimed to test the hypothesis that more language-specific models may be less sensitive to quantization for tasks in that language. As no clear trend emerged, we excluded this model from further experiments for the sake of conciseness.} Nonetheless, Qwen3 officially supports all the languages and dialects of WMT24++, except for the Canadian variant of French and Zulu. In contrast, Llama 3.1 and 3.3 officially support only English, French, German, Hindi, Italian, Portuguese, Spanish, and Thai.

\subsection{Quantization Methods and Hyperparameters}
We selected four quantization methods, primarily considering the accuracy reported in their respective papers and their popularity.\footnote{We considered GitHub stars, the number of quantized models available on the Hugging Face Hub, and support by widely-used inference frameworks such as vLLM, SGLang, and Transformers.}

\begin{itemize}
\item \textbf{AWQ}: Activation-aware Weight Quantization (AWQ) \cite{MLSYS2024_42a452cb} selects per-channel quantization scales based on activation magnitude distributions, explicitly reducing activation-sensitive quantization errors without requiring gradient-based calibration. We used AutoAWQ.\footnote{\url{https://github.com/casper-hansen/AutoAWQ}} For quantization, we used zero points with a group size of 128. AutoAWQ supports only 4-bit quantization.
\item \textbf{BnB}: bitsandbytes NormalFloat4 (NF4) \cite{10.5555/3666122.3666563} implements a learned, non-uniform 4-bit quantization scheme optimized for weight distributions that approximate normality, offering improved representational accuracy over uniform integer quantization. We used the \texttt{bitsandbytes}\footnote{\url{https://github.com/bitsandbytes-foundation/bitsandbytes}} integration with \texttt{Transformers},\footnote{\url{https://github.com/huggingface/transformers}} with nested quantization enabled. Note that bitsandbytes does not support 2-bit quantization.
\item \textbf{GGUF}: GGUF K-quantization with an importance matrix (imatrix) partitions weights into groups (size $K$) and applies group-wise quantization, guided by an importance matrix to quantize more critical weights more accurately. We used \texttt{llama.cpp}.\footnote{\url{https://github.com/ggml-org/llama.cpp}} The quantization strategies selected for this work were Q4\_K\_M and Q2\_K for 4-bit and 2-bit quantization, respectively. Imatrices were estimated using 20k random samples from WikiText,\footnote{\url{https://huggingface.co/datasets/Salesforce/wikitext}} with a context length of 512 and batch size of 512.
\item \textbf{AutoRound}: AutoRound \cite{cheng-etal-2024-optimize} employs differentiable rounding with a low-overhead signSGD-based optimization to fine-tune rounding decisions, minimizing task-specific loss directly rather than traditional reconstruction metrics. We used the AutoRound framework,\footnote{\url{https://github.com/intel/auto-round}} with 512 calibration samples, a maximum sequence length of 4096, and 512 optimization iterations. The group size was set to 32 and 128 for 2-bit and 4-bit quantization, respectively.

\end{itemize}

\subsection{Results}
Table \ref{tab:4bitFromEnglish} and Table \ref{tab:4bitIntoEnglish} confirm the expected first-order behavior: translation quality, measured by COMET, rises monotonically with model capacity and degrades after quantization. The smallest model (Qwen3-1.7B) loses up to $5$ COMET points, whereas Qwen3-32B and Llama-3.3-70B rarely lose more than one point. The language resource level moderates this effect: high-resource languages, such as Japanese and French, remain near their full-precision scores. In contrast, low-resource Indic languages and Zulu both start lower and incur the steepest drops. Among the PTQ algorithms, GGUF consistently yields the smallest loss, whereas AWQ and, in particular, BnB account for the largest degradations.

We also observe this strong inverse-performance law: the worse the baseline score of a language, the more it suffers from quantization.

Algorithmic robustness is not invariant to scale. BnB performs competitively at 8B but becomes the worst option at 70B.

\begin{table*}[]
    \centering
    \small
    \begin{tabular}{llcccccc}
         \toprule
 Model & PTQ (4-bit) & ja\_JP &  fr\_FR &  pl\_PL &  bn\_IN &  ml\_IN &  zu\_ZA \\         \midrule
         \midrule
         \multirow{5}{*}{Qwen3-1.7B}          & - &  80.4 & 74.2 & 64.7 & 50.1 & 35.9 & 27.3 \\
                                              & AWQ & 78.5 & 72.5 & 60.1 & 44.7 & 32.3 & 25.8 \\
                                              & BnB & 79.6 & 73.3 & 61.5 & 45.9 & 32.3 & 24.9 \\
                                              & GGUF & 79.7 & 74.3 & 64.2 & 48.5 & 35.8 & 26.0 \\
                                              & AutoRound & 78.2 & 73.3 & 62.2 & 45.8 & 31.3 & 27.7 \\
        \midrule
         \multirow{5}{*}{Qwen3-8B}          & - & 85.7 & 80.8 & 78.6 & 76.1 & 57.9 & 42.0 \\
                                            & AWQ & 85.0 & 80.1 & 77.6 & 74.1 & 55.4 & 41.9 \\
                                            & BnB & 85.4 & 80.0 & 77.6 & 74.0 & 55.1 & 43.0 \\
                                            & GGUF & 85.6 & 80.6 & 78.2 & 75.2 & 56.4 & 40.6 \\
                                            & AutoRound & 85.4 & 80.3 & 77.9 & 74.4 & 55.2 & 40.0 \\
                                                        \midrule

         \multirow{5}{*}{Llama-3.1-8B}          & - &81.6 & 77.7 & 78.0 & 75.6 & 57.9 & 45.5 \\ 
                                                & AWQ & 80.5 & 77.5 & 76.6 & 73.0 & 55.7 & 42.6 \\
                                                & BnB & 80.8 & 77.3 & 76.0 & 73.3 & 55.9 & 45.1 \\
                                                & GGUF & 81.2 & 77.2 & 77.7 & 74.6 & 55.0 & 44.0 \\
                                                & AutoRound & 80.3 & 77.4 & 77.0 & 74.1 & 54.6 & 44.9 \\
                                                        \midrule

         \multirow{5}{*}{Qwen3-32B}          & - & 87.1 & 81.7 & 81.8 & 80.0 & 65.4 & 54.9 \\
                                                & AWQ & 86.5 & 81.6 & 81.5 & 79.6 & 64.7 & 54.4 \\
                                                & BnB & 86.8 & 81.5 & 81.4 & 79.6 & 64.5 & 51.9 \\
                                                & GGUF & 87.2 & 81.7 & 81.7 & 79.8 & 65.4 & 54.1 \\
                                                & AutoRound & 86.8 & 81.7 & 81.8 & 79.6 & 64.9 & 52.8  \\
                                                        \midrule                                                

         \multirow{5}{*}{Llama-3.3-70B}          & - & 85.6 & 81.0 & 82.9 & 80.9 & 67.6 & 67.5 \\
                                                & AWQ & 85.2 & 80.8 & 82.7 & 80.5 & 67.0 & 66.8  \\
                                                & BnB & 84.8 & 80.5 & 81.9 & 79.3 & 65.1 & 65.5 \\
                                                & GGUF & 85.4 & 80.9 & 82.4 & 80.5 & 67.2 & 67.1 \\
                                                & AutoRound & 85.5 & 80.8 & 82.7 & 80.5 & 67.4 & 66.9  \\

         \bottomrule
    \end{tabular}
    \caption{COMET scores for translating from English with 4-bit quantization methods.}
    \label{tab:4bitFromEnglish}
\end{table*}

\begin{table*}[]
    \centering
    \small
    \begin{tabular}{llcccccc}
         \toprule
                  Model & PTQ (4-bit) & ja\_JP &  fr\_FR &  pl\_PL &  bn\_IN &  ml\_IN &  zu\_ZA \\

         \midrule
         \multirow{5}{*}{Qwen3-1.7B}         & - & 79.3 & 81.6 & 77.9 & 78.4 & 76.2 & 43.8 \\  
                                                & AWQ & 78.4 & 80.9 & 76.9 & 75.7 & 73.0 & 44.0 \\
                                                & BnB & 78.7 & 81.3 & 77.3 & 76.9 & 74.4 & 43.7 \\ 
                                                & GGUF & 79.2 & 81.4 & 77.6 & 78.0 & 75.4 & 43.5 \\
                                                & AutoRound & 79.0 & 81.0 & 77.0 & 76.8 & 74.1 & 42.9 \\
        \midrule
         \multirow{5}{*}{Qwen3-8B}         & - & 82.6 & 84.0 & 81.5 & 83.6 & 82.5 & 54.8  \\  
                                                & AWQ & 82.1 & 83.6 & 81.1 & 83.0 & 81.9 & 52.7 \\
                                                & BnB & 82.3 & 83.7 & 81.2 & 83.0 & 81.6 & 54.0 \\ 
                                                & GGUF & 82.5 & 84.0 & 81.2 & 83.4 & 82.4 & 53.4  \\
                                                & AutoRound & 82.3 & 83.6 & 81.2 & 82.9 & 81.6 & 52.5 \\
                                                        \midrule

         \multirow{5}{*}{Llama-3.1-8B}          & - & 79.3 & 81.6 & 79.1 & 72.2 & 71.8 & 58.4 \\  
                                                & AWQ &78.2 & 81.0 & 78.3 & 68.2 & 62.5 & 57.0 \\
                                                & BnB & 77.8 & 81.3 & 77.9 & 64.5 & 62.1 & 56.3\\ 
                                                & GGUF & 78.5 & 81.7 & 78.9 & 68.0 & 68.8 & 57.8 \\
                                                & AutoRound & 77.7 & 81.2 & 78.5 & 61.0 & 67.1 & 57.0 \\
                                                        \midrule
         \multirow{5}{*}{Qwen3-32B}         & - & 82.8 & 84.4 & 82.4 & 84.6 & 84.2 & 62.4  \\  
                                                & AWQ & 82.7 & 84.5 & 82.5 & 84.4 & 84.0 & 62.4  \\
                                                & BnB & 82.5 & 84.1 & 82.2 & 84.5 & 83.9 & 62.4 \\ 
                                                & GGUF & 82.8 & 84.4 & 82.5 & 84.1 & 83.9 & 62.2  \\
                                                & AutoRound & 82.8 & 84.5 & 82.4 & 84.3 & 84.0 & 61.9  \\
                                                        \midrule
         \multirow{5}{*}{Llama-3.3-70B}          & - & 80.9 & 83.3 & 81.1 & 82.6 & 82.2 & 66.9 \\  
                                                & AWQ & 80.8 & 83.3 & 81.2 & 82.7 & 82.2 & 66.7 \\
                                                & BnB & 80.0 & 82.5 & 80.2 & 78.2 & 79.4 & 60.7 \\ 
                                                & GGUF & 80.8 & 83.2 & 80.9 & 82.6 & 82.0 & 66.3 \\
                                                & AutoRound & 81.2 & 83.4 & 81.2 & 82.7 & 82.0 & 66.3 \\

         \bottomrule
    \end{tabular}
    \caption{COMET scores for translating into English with 4-bit quantization methods.}
    \label{tab:4bitIntoEnglish}
\end{table*}

Within every language pair, quantizing the weights down to two bits sharply reduces COMET scores (Table \ref{tab:2bitFromEnglish} and Table \ref{tab:2bitIntoEnglish}).

An intra-model asymmetry emerges when compression interacts with high-entropy Indic scripts. For instance, we have a ten-point cliff for Bengali while leaving the translation quality for Japanese and French almost intact: Qwen3-8B quantized to 2-bit with GGUF (English$\rightarrow$X), where the score change ranges from $-2$ COMET points on Japanese and French to $-17$ on Bengali and Malayalam.

BnB shows the sharpest intra-checkpoint contrast when the model is quantized and the source language is Indic. In the 4-bit tables, Llama-3.1-8B loses less than $2$ points on high-resource Latin or Kanji scripts (ja$\rightarrow$en: $-1.5$ COMET, fr$\rightarrow$en: $-0.3$ COMET) yet collapses for the Indic pairs (bn$\rightarrow$en: $-7.7$ COMET, ml$\rightarrow$en: $-9.7$ COMET). A milder but similar split appears for the larger Llama-3.3-70B model (pl$\rightarrow$en: $-0.9$ COMET, zu$\rightarrow$en: $-6.2$ COMET).

From these observations, we can conclude that the impact of quantization is language-dependent.

\begin{table*}[t]
    \centering
    \small
    \begin{tabular}{llcccccc}
         \toprule
 Model & PTQ (2-bit) & ja\_JP &  fr\_FR &  pl\_PL &  bn\_IN &  ml\_IN &  zu\_ZA \\         \midrule
         \multirow{3}{*}{Qwen3-1.7B}          & - &  80.4 & 74.2 & 64.7 & 50.1 & 35.9 & 27.3 \\  
                                               
                                                & GGUF & 69.4 & 66.2 & 48.2 & 30.8 & 23.7 & 31.2  \\
                                                & AutoRound & 33.1 & 34.8 & 32.9 & 28.6 & 24.5 & 27.1 \\

        \midrule
         \multirow{3}{*}{Qwen3-8B}          & - & 85.7 & 80.8 & 78.6 & 76.1 & 57.9 & 42.0\\  
                                               
                                                & GGUF & 83.7 & 79.2 & 72.5 & 59.9 & 40.6 & 37.7  \\
                                                & AutoRound & 47.0 & 63.9 & 49.7 & 34.4 & 31.9 & 32.4 \\
                                                        \midrule

         \multirow{3}{*}{Llama-3.1-8B}        & - & 81.6 & 77.7 & 78.0 & 75.6 & 57.9 & 45.5\\  
                                               
                                                & GGUF &  72.7 & 73.4 & 66.8 & 46.2 & 39.3 & 35.5 \\
                                                & AutoRound & 39.6 & 45.0 & 39.6 & 32.9 & 27.7 & 29.6\\ 
                                                        \midrule
         \multirow{3}{*}{Qwen3-32B}          & - &  87.1 & 81.7 & 81.8 & 80.0 & 65.4 & 54.9 \\  
                                               
                                                & GGUF & 86.0 & 80.5 & 78.1 & 74.7 & 56.1 & 44.0 \\
                                                & AutoRound & 81.1 & 75.0 & 68.0 & 54.5 & 37.3 & 39.0 \\
                                                  \midrule
         \multirow{3}{*}{Llama-3.3-70B}          & -  &   85.6 & 81.0 & 82.9 & 80.9 & 67.6 & 67.5\\  
                                               
                                                & GGUF & 84.2 & 79.8 & 81.0 & 78.9 & 63.2 & 58.7 \\
                                                & AutoRound & 81.7 & 78.7 & 75.5 & 68.4 & 54.0 & 41.8\\

         \bottomrule
    \end{tabular}
    \caption{COMET scores for translating from English with 2-bit quantization methods.}
    \label{tab:2bitFromEnglish}
\end{table*}

\begin{table*}[t]
    \centering
    \small
    \begin{tabular}{llcccccc}
         \toprule
 Model & PTQ (2-bit) & ja\_JP &  fr\_FR &  pl\_PL &  bn\_IN &  ml\_IN &  zu\_ZA \\         \midrule
         \midrule
                \multirow{3}{*}{Qwen3-1.7B}          & - & 79.3 & 81.6 & 77.9 & 78.4 & 76.2 & 43.8 \\  
                                               
                                                & GGUF & 71.9 & 77.8 & 70.9 & 59.6 & 56.2 & 41.6 \\
                                                & AutoRound & 43.0 & 43.6 & 37.3 & 42.1 & 38.9 & 35.5 \\
                                                        \midrule
                \multirow{3}{*}{Qwen3-8B}          & - & 82.6 & 84.0 & 81.5 & 83.6 & 82.5 & 54.8 \\  
                                               
                                                & GGUF & 80.9 & 82.8 & 79.7 & 81.3 & 79.8 & 44.0  \\
                                                & AutoRound & 71.4 & 74.3 & 70.1 & 64.6 & 58.7 & 38.0 \\
                                                        \midrule                                                        

         \multirow{3}{*}{Llama-3.1-8B}        & - & 79.3 & 81.6 & 79.1 & 72.2 & 71.8 & 58.4 \\  
                                               
                                                & GGUF & 76.3 & 79.9 & 76.0 & 70.2 & 68.1 & 47.6 \\
                                                & AutoRound & 55.8 & 54.3 & 50.8 & 37.5 & 43.0 & 38.3\\ 
                                                        \midrule
         \multirow{3}{*}{Qwen3-32B}          & - &  82.8 & 84.4 & 82.4 & 84.6 & 84.2 & 62.4\\  
                                               
                                                & GGUF & 81.0 & 83.1 & 78.8 & 79.9 & 79.5 & 57.1 \\
                                                & AutoRound & 80.7 & 82.1 & 79.7 & 81.7 & 79.8 & 51.1 \\
                                                  \midrule
         \multirow{3}{*}{Llama-3.3-70B}          & -  & 80.9 & 83.3 & 81.1 & 82.6 & 82.2 & 66.9\\  
                                               
                                                & GGUF & 80.1 & 82.1 & 80.3 & 80.5 & 80.7 & 61.8 \\
                                                & AutoRound &  79.9 & 81.5 & 79.4 & 79.6 & 80.1 & 59.8\\

         \bottomrule
    \end{tabular}
    \caption{COMET scores for translating into English with 2-bit quantization methods.}
    \label{tab:2bitIntoEnglish}
\end{table*}

\section{Inference Hyperparameters: Temperature and Nucleus Sampling}

We quantified the joint impact of weight quantization and sampling hyperparameters on the Llama-3.1-8B-Instruct model when translating Japanese, French, and Bengali into English. Results are reported with COMET (Table~\ref{tab:cometllama}) and chrF (Table~\ref{tab:chrfllama}).

For each model variant we swept five values for temperature $T\in\{0.0,0.2,0.4,0.6,0.8\}$ and two nucleus thresholds $\text{top-}p\in\{1.0,0.95\}$. The reference (``Original'') model was compared with two GGUF quantizations: Q4\_K\_M (4-bit) and Q2\_K (2-bit). All other decoding parameters remained at their default values.

\begin{itemize}
    \item \textbf{Sampling temperature dominates top-$p$.}
For all weight precisions, raising $T$ monotonically lowers quality.
Moving from deterministic decoding ($T{=}0$) to $T{=}0.8$
degrades COMET by roughly $1.3$--$1.6$ points in Japanese and French
and by $1.4$--$1.5$ points in Bengali. In contrast, tightening nucleus sampling from $1.0$ to $0.95$
changes COMET by at most $\pm0.2$ points and chrF by
$\pm0.3$ points, well within typical scoring variance.

\item \textbf{No strong interaction between quantization and decoding hyperparameters.}
The relative ranking of weight formats is stable across
the five temperature and two top-$p$ values.

\end{itemize}

\begin{table}[t]
    \centering
    \small
    \begin{tabular}{lcccc}
\toprule Model & $T$/top-$p$ & ja\_JP & fr\_FR & bn\_IN\\ \midrule
\multirow{9}{*}{Original}& 0.0/1.0 & 79.2 & 81.7 & 72.4 \\
&0.2/1.0 & 79.1 & 81.7 & 72.2 \\
&0.4/1.0 & 79.0 & 81.8 & 71.9 \\
&0.6/1.0 & 78.6 & 81.3 & 71.5 \\
&0.8/1.0 & 77.9 & 81.2 & 70.9 \\ \cmidrule{2-5}
&0.2/0.95 & 79.1 & 81.8 & 72.0 \\
&0.4/0.95 & 78.9 & 81.6 & 72.0 \\
&0.6/0.95 & 78.6 & 81.5 & 71.8 \\
&0.8/0.95 & 78.5 & 81.2 & 71.3 \\

\midrule
\multirow{9}{*}{GGUF Q4\_K\_M} &0.0/1.0 & 78.7 & 81.5 & 67.5 \\
&0.2/1.0 & 78.7 & 81.5 & 67.3 \\
&0.4/1.0 & 78.4 & 81.6 & 67.3 \\
&0.6/1.0 & 78.3 & 81.4 & 66.9 \\
&0.8/1.0 & 77.6 & 81.1 & 66.1 \\ \cmidrule{2-5}
&0.2/0.95 & 78.7 & 81.5 & 67.4 \\
&0.4/0.95 & 78.6 & 81.5 & 67.3 \\
&0.6/0.95 & 78.5 & 81.4 & 67.0 \\
&0.8/0.95 & 77.8 & 81.1 & 66.6 \\
\midrule
\multirow{9}{*}{GGUF Q2\_K} &0.0/1.0 & 76.4 & 80.1 & 70.0 \\
&0.2/1.0 & 76.3 & 80.0 & 70.5 \\
&0.4/1.0 & 76.1 & 79.6 & 69.9 \\

&0.6/1.0 & 75.4 & 79.4 & 69.2 \\
&0.8/1.0 & 74.3 & 79.3 & 67.8 \\ \cmidrule{2-5}
&0.2/0.95 & 76.1 & 80.0 & 70.3 \\
&0.4/0.95 & 76.2 & 79.9 & 69.9 \\
&0.6/0.95 & 75.8 & 79.5 & 69.9 \\
&0.8/0.95 & 75.2 & 79.4 & 69.0 \\
\bottomrule
    \end{tabular}
    \caption{COMET scores with Llama-3.1-8B-Instruct translating into English while varying temperature and top-$p$.}
    \label{tab:cometllama}
\end{table}

\begin{table}[t]
    \centering
    \small
    \begin{tabular}{lcccc}
\toprule Model & $T$/top-$p$ & ja\_JP & fr\_FR & bn\_IN\\ \midrule
\multirow{9}{*}{Original}&0.0/1.0 & 45.5 & 58.93 & 32.61 \\
&0.2/1.0 & 45.3 & 59.0 & 32.3 \\
&0.4/1.0 & 45.0 & 58.8 & 31.9 \\
&0.6/1.0 & 44.3 & 58.4 & 31.3 \\
&0.8/1.0 & 44.0 & 58.0 & 31.4 \\ \cmidrule{2-5}
&0.2/0.95 & 44.8 & 58.0 & 31.4 \\
&0.4/0.95 & 45.2 & 58.8 & 31.9 \\
&0.6/0.95 & 44.2 & 58.4 & 31.6 \\
&0.8/0.95 & 44.1 & 58.2 & 31.4 \\

\midrule
\multirow{9}{*}{GGUF Q4\_K\_M} &0.0/1.0 & 43.8 & 59.0 & 18.7 \\
&0.2/1.0 & 44.0 & 58.7 & 18.8 \\
&0.4/1.0 & 43.8 & 58.8 & 18.9 \\
&0.6/1.0 & 43.4 & 58.6 & 19.6 \\
&0.8/1.0 & 42.3 & 58.0 & 19.2 \\ \cmidrule{2-5}
&0.2/0.95 & 43.9 & 59.0 & 19.1 \\
&0.4/0.95 & 43.7 & 58.2 & 19.2 \\
&0.6/0.95 & 43.6 & 58.6 & 20.1 \\
&0.8/0.95 & 42.6 & 57.9 & 19.8 \\
\midrule
\multirow{9}{*}{GGUF Q2\_K} &0.0/1.0 & 41.8 & 56.9 & 34.3 \\
&0.2/1.0 & 41.5 & 56.6 & 34.7 \\
&0.4/1.0 & 41.4 & 56.3 & 33.6 \\
&0.6/1.0 & 41.2 & 55.2 & 34.2 \\
&0.8/1.0 & 38.9 & 55.3 & 33.7 \\ \cmidrule{2-5}
&0.2/0.95 & 41.6 & 56.8 & 33.6 \\
&0.4/0.95 & 41.6 & 56.5 & 33.9 \\
&0.6/0.95 & 41.2 & 56.5 & 34.8 \\
&0.8/0.95 & 40.6 & 55.9 & 34.8 \\
\bottomrule
    \end{tabular}
    \caption{chrF scores with Llama-3.1-8B-Instruct translating into English while varying temperature and top-$p$.}
    \label{tab:chrfllama}
\end{table}

\section{Impact of the Calibration Step}
We evaluated whether calibrating quantization on the target language improves translation performance into and from that language. Specifically, we compared the Llama-3.1-8B-Instruct model quantized with GGUF Q4\_K\_M and Q2\_K formats, where quantization statistics (imatrix) were collected either on English or Bengali data.\footnote{We randomly sampled 10k tokens from the dataset HuggingFaceFW/fineweb-2 (\url{https://huggingface.co/datasets/HuggingFaceFW/fineweb-2}) in English and Bengali to create calibration datasets in these languages.} We evaluated four translation directions: English into French and Bengali (Table~\ref{tab:fromcalib}) and French and Bengali into English (Table~\ref{tab:intocalib}). COMET scores are reported for all settings.

\paragraph{Calibration language has minimal impact in 4-bit quantization.}
For Q4\_K\_M, calibrating on Bengali versus English yields negligible differences in COMET scores. In both translation directions and for both evaluation languages (French and Bengali); changes are within $0.2$ points. This suggests that 4-bit quantization preserves enough representational flexibility that the specific language used for calibration has limited influence on final quality, at least in models of this size and under these conditions.

\paragraph{2-bit quantization fairly benefits from language-matched calibration.}
In contrast, for Q2\_K quantization, using Bengali calibration improves COMET scores for Bengali translation in both directions. For translation into English, Bengali calibration leads to a $0.8$ point gain over English calibration ($71.5$ vs.\ $72.3$). For translation from English, the improvement is more pronounced: $48.0$ vs.\ $44.9$ ($+3.1$). This suggests that calibrating on the target language can partially mitigate the degradation caused by aggressive (2-bit) quantization, particularly in low-resource settings where representational margins are narrower.

\paragraph{No benefit for unrelated languages.}
For French, which is not involved in calibration, the COMET scores remain nearly identical regardless of the calibration language. This further supports the idea that the benefits of language-matched calibration are specific and do not generalize to unrelated language pairs. French scores even slightly improve under Q2\_K when calibrated on Bengali.

\begin{table}[t]
    \centering
    \small
    \begin{tabular}{lccc}
\toprule Model & Calib. Language & fr\_FR & bn\_IN\\ \midrule
Original & - &   77.7 &  75.6  \\
\midrule
\multirow{2}{*}{Q4\_K\_M} & English & 77.2 & 75.1 \\
 & Bengali & 77.3 & 75.1 \\
\midrule
\multirow{2}{*}{Q2\_K} & English & 73.8 & 44.9  \\
 & Bengali & 73.5 &  48.0\\
\bottomrule
    \end{tabular}
    \caption{COMET scores for Llama-3.1-8B-Instruct translating from English with a quantized model calibrated with either English or Bengali datasets.}
    \label{tab:fromcalib}
\end{table}

\begin{table}[t]
    \centering
    \small
    \begin{tabular}{lccc}
\toprule Model & Calib. Language & fr\_FR & bn\_IN\\ \midrule
Original & - & 81.7 & 72.4 \\
\midrule
\multirow{2}{*}{Q4\_K\_M} & English &  81.6& 63.9  \\
 & Bengali & 81.6 &  65.8\\
\midrule
\multirow{2}{*}{Q2\_K} & English &79.8 & 72.3 \\
 & Bengali & 80.2 &  71.5\\
\bottomrule
    \end{tabular}
    \caption{COMET scores for Llama-3.1-8B-Instruct translating into English with a quantized model calibrated with either English or Bengali datasets.}
    \label{tab:intocalib}
\end{table}

\section{Limitations}
\label{sec:limitations}

Although our experiments span five model sizes, four quantization algorithms, two bit-widths, and 55 languages, the findings remain subject to several nuanced constraints that limit the interpretability and generalizability of the results.

\paragraph{Metric Fragility Beyond Point Estimates.}
Our conclusions about translation quality are based on single-run COMET scores. Since post-training quantization (PTQ) introduces stochastic rounding noise, results are not bit-exact across different library versions or GPU kernels. As a result, reproducing the exact rankings or scores using an alternative implementation of the same PTQ recipe cannot be guaranteed.

\paragraph{Dataset-Specific Calibration Bias.}
The calibration data used for imatrix estimation and AutoRound consists of generic Wikipedia or FineWeb snippets, rather than in-domain parallel corpora. This biases the learned scale factors toward short, well-formed sentences and Latin-based scripts. Consequently, low-resource languages with divergent grapheme inventories (e.g., Malayalam and Amharic) may receive suboptimal quantization ranges, even at the same bit-width.

\paragraph{Prompt and Decoding Entanglement.}
We adopt a fixed direct-translation prompt and greedy decoding across all evaluations. However, quantization noise interacts with all components of the decoding pipeline. Early experiments show that beam search magnifies minor rounding errors into larger syntactic deviations, while greedy decoding (temperature of 0.0) can obscure some of these artifacts. Since we do not perform a full sweep over prompts and decoding strategies, our results reflect only a narrow region of the search space; real-world deployments may exhibit different sensitivity profiles.

In a nutshell, while this study offers the first large-scale analysis of multilingual MT under aggressive PTQ, its conclusions are conditioned on a limited subset of the broader design space. Future research should incorporate activation and KV cache quantization, domain-specific calibration, fine-grained linguistic error analysis, and controlled human evaluations to validate trends observed with automatic metrics.

\section{Conclusion}
This work provides an analysis of how post-training quantization affects MT performance in multilingual LLMs. We show that while 4-bit quantization is generally reliable, especially for larger models, its effects are not uniform across languages. Low-resource and structurally diverse languages are disproportionately impacted, sometimes losing over $10$ COMET points, while preserving translation quality for high-resource languages. Our findings also underscore that quantization algorithms are not equally robust: GGUF offers the most consistent results across model sizes and bitwidths, while BnB performs well only for small models and degrades translation quality at larger scales. Additionally, calibration on the target language can mitigate performance loss in 2-bit settings. Altogether, these results highlight the importance of multilingual-aware quantization benchmarks and tuning strategies for real-world deployment. Future work should explore the joint impact of activation and KV cache quantization, and evaluate new state-of-the-art quantization formats or data types recently introduced, such as MXFP4 and NVFP4.
\bibliography{custom}

\appendix

\section{COMET, BLEU, and chrF Scores}
The scores for all the models and translation directions can be found in this dataset: bnjmnmarie/wmt24pp-qtranslated-scores.\footnote{\url{https://huggingface.co/datasets/bnjmnmarie/wmt24pp-qtranslated-scores}}

We used SacreBLEU \cite{post-2018-call}\footnote{\url{https://github.com/mjpost/sacrebleu}} to compute the chrF\footnote{Signature: nrefs:1|case:mixed|eff:yes|nc:6|nw:0|space:no|version:2.5.1} scores and BLEU\footnote{Signature: nrefs:1|case:mixed|eff:no|tok:13a|smooth:exp|version:2.5.1} scores.

\end{document}